\pdfoutput=1

\documentclass[11pt]{article}

\usepackage[]{EMNLP2022}

\usepackage{times}
\usepackage{latexsym}

\usepackage[T1]{fontenc}

\usepackage[utf8]{inputenc}

\usepackage{microtype}

\usepackage{inconsolata}

\usepackage{latexsym}
\usepackage{tikz}
\usepackage{pgfplots}
\usepackage{capt-of}
\usepackage{scalefnt}
\usepackage{subcaption}
\usepackage{comment}
\usepackage{float}
\usetikzlibrary{patterns}
\pgfplotsset{compat=1.17}
\usepackage{amsmath}
\usepackage{amssymb}
\usepackage{mathtools}
\usepackage{amsthm}
\usepackage[capitalize,noabbrev]{cleveref}

\title{You Only Need One Model for Open-domain Question Answering}

\author{Haejun Lee$^\clubsuit$ \qquad Akhil Kedia$^\clubsuit$ \qquad Jongwon Lee$^\diamondsuit$* \qquad Ashwin Paranjape$^{\spadesuit}$\\
  {\bf Christopher D. Manning}$^{\spadesuit}$ \qquad {\bf Kyoung-Gu Woo}$^\heartsuit$*\\
 $\clubsuit$ Samsung Research \qquad $\diamondsuit$ Samsung Electronics\\
 $\spadesuit$ Stanford University \qquad $\heartsuit$ Growdle Corporation \\

  \texttt{\{haejun82.lee, akhil.kedia, 	jay722.lee\}@samsung.com} \\
  \texttt{\{ashwinp, manning\}@cs.stanford.edu, epigramwoo@growdle.com} }

\begin{document}
\maketitle

\newif\ifaclfinal
\aclfinaltrue
\ifaclfinal
\renewcommand*{\thefootnote}{\fnsymbol{footnote}}
\setcounter{footnote}{1}
\footnotetext{Work done while at Samsung Research}
\renewcommand*{\thefootnote}{\arabic{footnote}}
\setcounter{footnote}{0}
\fi

\begin{abstract}
Recent approaches to Open-domain Question Answering refer to an external knowledge base using a retriever model, optionally rerank passages with a separate reranker model and generate an answer using another reader model. Despite performing related tasks, the models have separate parameters and are weakly-coupled during training. We propose casting the retriever and the reranker as internal passage-wise attention mechanisms applied sequentially within the transformer architecture and feeding computed representations to the reader, with the hidden representations progressively refined at each stage. This allows us to use a single question answering model trained end-to-end, which is a more efficient use of model capacity and also leads to better gradient flow. We present a pre-training method to effectively train this architecture and evaluate our model on the Natural Questions and TriviaQA open datasets. For a fixed parameter budget, our model outperforms the previous state-of-the-art model by 1.0 and 0.7 exact match scores.
\end{abstract}
\section{Introduction}

Open-domain Question Answering (Open QA) is a knowledge-intensive task that finds the answer for the given question from a large-scale knowledge corpus that can easily surpass millions of documents. Thus, how to store and refer to the knowledge at such scales is important in terms of both performance and scalability for Open QA systems. Traditional systems rely on information retrieval engines such as Lucene. These score the relevance of knowledge to a given query by lexical overlaps between them in a sparse representation space based on TF-IDF or BM25 \cite{chen2017reading,wang2018evidence,yang-etal-2019-end-end}. However, recent advances in neural language modeling have enabled two new lines of approach; 1) referring to internal knowledge parameterized in the model \cite{gpt3,petroni2019language,roberts-etal-2020-much}, and 2) referring to external knowledge retrieved by matching query and knowledge in dense representation spaces \cite{dpr,orqa,realm,rag,fid}.

Despite the simplicity of the approach, parametric models have limitations such as a large number of model parameters that require large compute for both training and inference and non-expandable knowledge without re-training. Their implicit knowledge reference also makes it hard to find supporting knowledge and often results in hallucinations \cite{tread}. The current dense retrieval models have advantages over parametric models on these issues \cite{dpr,realm,rag}. But most retrieval models only have a weak coupling between and separate parameters for the reader, reranker (if any), and retriever that limits these models from achieving optimal end-to-end training and efficient use of the total model capacity. 

\begin{figure*}[hbt]
\includegraphics[width=\linewidth]{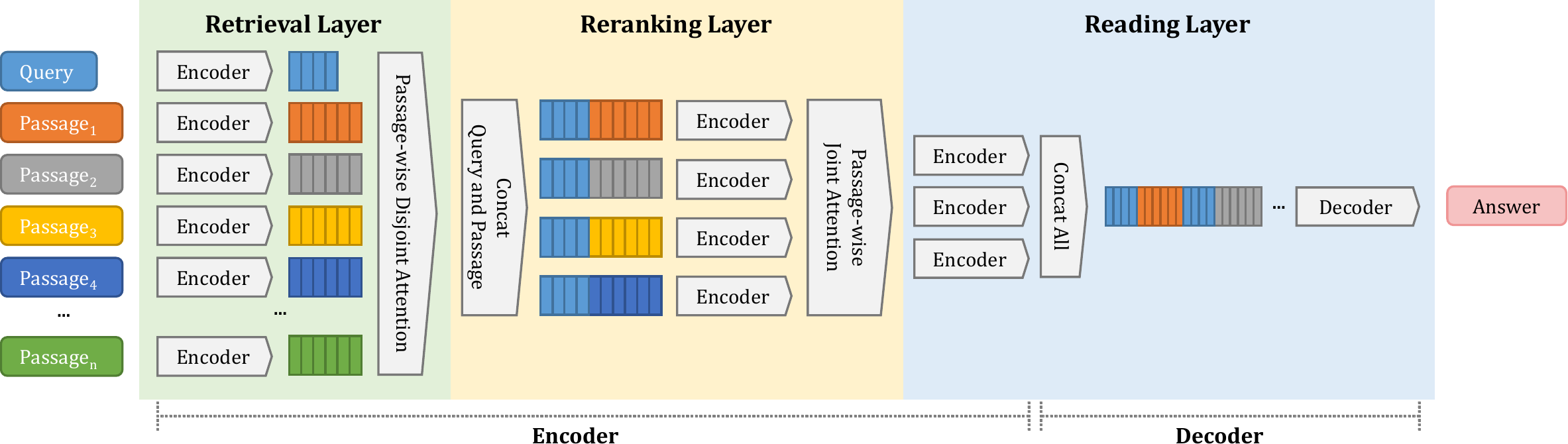} 
\caption{The overall architecture of our proposed model YONO.}
\label{rian_architecture}
\end{figure*}

In this paper, we propose a single language model YONO (You Only Need One model) that can refer to external knowledge via its internal attention functions, which are trainable in a fully end-to-end manner. We achieve this by generalizing the retrieval and reranking as internal passage-wise attentions. At the lower retrieval layers, the query and passages are separately encoded allowing pre-computation of all the passage representations. Then passage-wise hard attention is applied to retrieve initial relevant passages from the entire knowledge base. While it would be optimal to retrieve passages based on cross-attention between the query and all the passages \cite{colbert}, it is computationally intractable. Hence, we approximate this attention by a passage-wise hard-attention layer using decoupled query and passage representations. The representations of the initial relevant passages are further encoded jointly with the query representation to compute more expressive coupled representations. These are used to select only the more relevant passages using another passage-wise hard-attention in the reranking layer. The representations of the final set of passages are then encoded by transformer encoders for deeper representations that are fused in the decoder to generate the answers. 

We train this architecture fully end-to-end by self-supervised pre-training and weakly supervised fine-tuning without passage labels. 

Our contributions are twofold;

\begin{itemize}
    \item A single model that generalizes retrieval, reranking, and reading as internal attention functions. We show that this model trained end-to-end significantly improves the retrieval performance by leveraging a training signal from the answer generation decoder to allow better gradient flow across the whole model in \cref{reader_loss}. It also achieves better utilization of the model parameters, outperforming a stand-alone reader with the same number of parameters by 7.1\% and 3.2\% on NQ and TQA respectively as shown in \cref{shared_rep}.
    \item A method to train this architecture in a fully end-to-end manner. We show our pre-training method requires 51.5\% fewer pre-training tokens compare to the previous the state-of-the-art approach in \cref{pre-train-overhead}.
\end{itemize}

\section{YONO Architecture}

As depicted in \cref{rian_architecture}, we propose a single encoder-decoder language model architecture consisting of 3 components: Retrieval Layer, Reranking Layer, and Reading Layer. 

\subsection{Retrieval Layer}

This first layer retrieves the top-N relevant passages for a given query from the knowledge corpus using query and passage representations independently encoded with the first $K$ transformer encoder layers. The query is encoded with `\textit{query}:' prefix while passages are encoded with `\textit{title}:' and `\textit{context}:' prefixes following previous approaches \cite{fid,emdr2}. 

\paragraph{Passage-wise Disjoint Attention:} Let $q_0$ and $P_0$ be the first tokens' representations of the query and all the passages respectively encoded independently. The disjoint attention scores are calculated by the scaled dot-product attention scores \cite{NIPS2017_3f5ee243} between $q_0$ and $P_0$ as:
\begin{align}\label{attention_score}
Q=L&ayerNorm(q_0W_q)\nonumber\\
K=L&ayerNorm(P_{0}W_p)\nonumber\\
score_{disjoint}&(q,P)=\sigma({QK^T}/{\sqrt{d_k}})
\end{align}
where $W_q, W_p \in \mathrm{R^{d \times d}}$ are learned linear projections and $1/\sqrt{d_k}$ is the scaling factor following \citet{NIPS2017_3f5ee243}. 

The top-N relevant passages $P^R$ with the highest  $score_{disjoint}$ for a given query are selected and passed to the next layer. In practice, we retrieve top-N passages by indexing the pre-computed passage representations $P_0$ using Maximum Inner Product Search tools (MIPS) such as FAISS \citep{faiss}. During training, the index is iteratively refreshed by the most recent model's representation following other neural retrieval approaches \cite{dpr,orqa,realm,fid}.

\subsection{Reranking Layer}

We further narrow down the retrieved passages by applying an additional passage-wise attention based on more expressive representations from the joint encoding of query and passages. 

\paragraph{Passage-wise Joint Attention:} We concatenate query and retrieved passage representations and encode them with cross-attention for more expressive representations using the next $L$ transformer encoder layers. Let $h^n$ be the encoded representations of query and the $n^{th}$ passage and $H_0$ be the first token's representations of all encoded representations. We apply the second passage-wise attention based on $score_{joint}(q,P^R)$, obtained from $H_0$:
\begin{align}\label{cross_attention_score}
h^n&=Transformer(q\oplus p^n)\\
H^0&=[h_0^0, h_0^1, h_0^2, ..., h_0^{N-1}]\nonumber\\
score_{joint}&(q,P^R)=\sigma(LayerNorm(H_0)W_{qp})\nonumber
\end{align}
where $\oplus$ is the concatenation operation and $W_{qp} \in \mathrm{R^{d \times 1}} $ is a learnt vector.

\subsection{Reading Layer}
After the retrieval and reranking layers, the final representations are fed to the reading layer. These are further encoded using the remaining transformer encoder layers and fused in the decoder for multi-passage reading, following the approach of FiD \cite{fid}.

\section{Training YONO}
\subsection{Training Objective}
\label{training_objective}
The whole model is always trained end-to-end by leveraging a training signal from the final answer generation. Due to non-differentiability of the passage-wise attentions, we combine additional losses $\mathcal{L}_{retrieval}$ and $\mathcal{L}_{reranking}$ to the answer generation $\mathcal{L}_{reading}$ as below:
\begin{equation}
\label{final_loss}
\mathrm{\mathcal{L}}=\mathcal{L}_{retrieval}+\mathcal{L}_{reranking}+\mathcal{L}_{reading}
\end{equation}

\paragraph{Retrieval and Reranking Loss:} The passage-wise attention scores $S_{retrieval}$ and $S_{reranking}$ for retrieval and reranking layers are calculated by \cref{attention_score} and (\ref{cross_attention_score}) using retrieved passages $P^R$ and in-batch negative passages $P^N$ as:
\begin{align}\label{scores}
S_{retrieval}=score_{disjoint}&(q,P^R \cup P^N) \nonumber\\
S_{reranking}=score_{joint}&(q,P^R)
\end{align}
In-batch negative passages $P^N$ are used to expand $S_{retrieval}$ for more contrastive training signals. In-batch negatives not used for $S_{reranking}$ because the joint representation is only calculated for the retrieved passages, not the in-batch negatives. 

These attention scores are not differentiable by the reader's generation loss because they are only used to select top-N passages at retrieval and reranking layers but not directly used in the answer generation. Instead, we train $score_{retrieval}$ and $score_{reranking}$ to approximate the target scores, which following previous work \cite{fidkd} are derived by accumulating the decoder's attention scores across decoder layers and attention heads over all encoded passage tokens as: 
\begin{align}\label{soft_attention_score}
score_{decoder}&(P)=\\
\sum^{N_l}_{l=0}\sum^{N_h}_{h=0}&\sum^{N_t^p}_{t_p=0}{\frac{SG(att_{dec}(0,l,h,t_p))}{N_lN_hN_t^p}}~\vert ~ p \in P\nonumber
\end{align}
where $att_{dec}$ is decoder attention matrices toward encoded outputs, $0$ is an output token index, $N_l$ is the number of layers, $N_h$ is the number of attention heads, $N^p_t$ is the number of tokens in a given passage, and $SG$ is a stop gradient function. The gradient flow back to the decoder's attention scores is blocked by $SG$ to train the decoder by $\mathcal{L}_{reading}$ only.

Using this scoring function, we get target scores $T_{retrieval}$ and $T_{reranking}$. The scores of in-batch negative passages $P^N$ are set to 0.
\begin{align}\label{ground_truth}
T_{retrieval}&=score_{decoder}(P^R) \oplus (0~\vert ~ p \in P^N) \nonumber\\
T_{reranking}&=score_{decoder}(P^R) 
\end{align}

Finally, the losses for training passage-wise attention scores are obtained by KL-Divergence between target scores $T$ and attention scores $S$ as:
\begin{align}\label{hard_attention_loss}
\mathrm{\mathcal{L}}_{retrieval}=D_{KL}(T_{retrieval} &\Vert S_{retrieval}) \nonumber\\
\mathrm{\mathcal{L}}_{reranking}=D_{KL}(T_{reranking} &\Vert S_{reranking})
\end{align}

In addition, we add a constant penalty $\gamma$ to $score_{retrieval}$ of in-batch negative passages $P^N$ before applying a softmax of \cref{attention_score}. This inductive bias enforces the model to further decrease scores of the random negative passages lower than the lowest score of the retrieved passages.

\paragraph{Reading Loss:} We use a conventional auto-regressive language modeling loss for generating an answer $a$ given a query $q$ and retrieved passages $P^R$:
\begin{equation}\label{reading_loss}
\mathrm{\mathcal{L}}_{reading}=-\log{\prod_{t=1}^{T_A}{p(a_t~\vert ~ a_{<t}, q, P^R)}}
\end{equation}

\subsection{Pre-training Corpus}
We first pre-train our model to adapt the pre-trained encoder-decoder architecture to the YONO architecture and provide initial retrieval performance for fine-tuning without passage labels. Inverse Cloze Task (ICT) \cite{orqa} and Masked Salient Span (MSS) \cite{roberts-etal-2020-much,realm} are widely used tasks for pre-training. ICT uses `\textit{input-passage}' pairs that have explicit supervision for training passage-wise attention, but has no supervision for the answer generation. On the other hand, MSS trains the model by `\textit{input-output}' pairs that have strong supervision for the answer generation, but no supervision for the retriever, requiring additional warm-up training for retrieval such as ICT. Thus, these tasks are sequentially applied to pre-train the pipeline models to overcome their limitations \cite{realm,emdr2}. However, we train our single model architecture with retrieval, reranking, and reading layers at the same time using triples of `\textit{input-passage-output}' for pre-training. To provide such supervisions, we extend a masked salient span task with explicit passage labels.

\paragraph{Masked Salient Span with Passage Labels (MSS-P):}  We first pick one named entity and mask all instances of this entity from the sentence. We explicitly add a ground truth passage that contains the masked named entity from 2 previous and next passages except its original passage. We refine the data by simple heuristics (using only pairs of sentence and target passage that contain at least 1 common named entity other than the masked span, and selecting the target passage with the highest number of common named entities when there are multiple passages containing the masked span). In this way, we generate 53M triples from Wikipedia passages in total.

\subsection{Training Procedure}
\label{train_procedure}

The model is pre-trained and fine-tuned iteratively to refresh retrieved passages for a better approximation of the distribution over all the passages. 

We start the first pre-training iteration using the initial pre-training data, extracted by the MSS-P method that has one ground-truth passage for each query sentence. Note that with one positive passage per query, $\mathcal{L}_{retrieval}$ is equivalent to the negative log-likelihood loss of predicting the positive passage along with negative passages. However, $\mathcal{L}_{reranking}$ is 0 at this pre-training iteration and does not yield any training signal because it can only learn from contrasting multiple retrieved passages.

From the second iteration, the model is trained with 100 passages fetched from the retrieval layer. We add the original ground truth passage to the retrieved passages to ensure that the model learns to refer to the knowledge instead of implicitly memorizing the answer in its internal parameters. We do not filter more passages at the reranking layer during training to compute $score_{decoder}$ of \cref{soft_attention_score} to allow the reader to learn from the maximum number of passages. We pre-train the model for several iterations until the performance of the retrieval converges based on the recall metric. 

After the pre-training, the model is then fine-tuned following the same procedure as that after the first iteration. To prevent an over-fitting of the reader due to the limited size of the fine-tuning data, we simply re-initialize the model with the pre-trained YONO model after retrieval performance converges following \citet{fidkd}. Note that the model is fine-tuned by only weak-supervision of question-answer pairs without gold passages. 

\section{Experiments}
\label{experiments}

\begin{table*}[hbt]
\begin{center}
\begin{small}
\setlength{\tabcolsep}{0.48em}

\begin{tabular}{lccccccccccc}
\hline
 & \textbf{\scalebox{.9}[1.0]{Passage}} & \textbf{Aug.} & \textbf{Model} & \multicolumn{3}{c}{\textbf{Natural Questions}} & & \multicolumn{3}{c}{\textbf{TriviaQA}} \\ 
         \cline{5-7} \cline{9-11}  
\textbf{Model} & \textbf{Label} & \textbf{data} & \textbf{\scalebox{.9}[1.0]{\# Params}}  &\textbf{R@5} & \textbf{R@20} & \textbf{R@100}  & & \textbf{R@5} & \textbf{R@20} & \textbf{R@100}\\ 
\hline
BM25 \cite{gar} & & & - & 43.6 & 62.9 & 78.1 & & 67.7 & 77.3 & 83.9 \\
\hline
DPR \cite{dpr} &  $\surd$ &  & 220M & 68.1 & 80.0 & 85.9  & & - & 79.4 & 85.0 \\
DPR$^{new}$ \cite{dpr} & $\surd$ & & 220M & 72.2 & 81.3 & 87.3  & & - & - & - \\
GAR$^+$  \cite{gar} & $\surd$ & $\surd$ & 220M & 70.7 & 81.6 & 88.9  & & 76.0 & 82.1 & 86.6 \\
PAIR \cite{pair}& $\surd$ & $\surd$ & 220M & 74.9 & 84.0 & 89.1  & & - & - & - \\
coCondenser \cite{gao2021condenser} & $\surd$ & & 220M & \textbf{75.8} & 84.3 & 89.0  & & 76.8 & 83.2 & 87.3 \\
DPR-PAQ \cite{dprpaq} & $\surd$ & $\surd$ & 220M & 74.2 & 84.0 & 89.2 & & - & - & -\\
ANCE \cite{ance} & $\surd$ & & 220M & - & 81.9 & 87.5   & & - & 80.3 & 85.2 \\
E2NR \cite{e2nr} & & & 220M & 75.0 & 84.0 & 89.2 & & 76.8 & 83.1 & 87.0 \\
R2-D2$_{Retrieval}$ \cite{r2d2}  & $\surd$ &  & 220M & 68.6 & 80.6 & 86.7  & & 69.8 & 78.9 & 84.7 \\
FiD-KD \cite{fidkd} & $\surd$ & & 220M & 73.8 & 84.3 & 89.3 & & \textbf{77.0} & \textbf{83.6} & \textbf{87.7} \\
\hline
\multicolumn{5}{l}{\textit{\textbf{Larger models} }} \\
E2NR \cite{e2nr} & & & 660M & 76.2 & 84.8 & 89.8 & & 78.7 & 84.1 & 87.8 \\
DPR-PAQ \cite{dprpaq}  & $\surd$ & $\surd$ & 660M & 76.9 & 84.7 & 89.2  & & - & - & - \\

\hline
\textbf{YONO$_{Retrieval}$} & & & 165M & 75.3 & \textbf{85.2} & \textbf{90.2} & & 76.8 & 83.5 & 87.4  \\
\hline
\multicolumn{5}{l}{\textit{\textbf{Reranker models}}} \\
GAR$^+$-BART \cite{rider}  & $\surd$ & & 406M & 73.5 & 82.2 & -  & & - & - & - \\
GAR$^+$-RIDER \cite{rider}  & $\surd$ & & 110M & 75.2 & 83.2 & 88.9  & & 77.9 & 82.8 & 85.7 \\
R2-D2\scalebox{.7}[1.0]{$_{Reranking200}$} \cite{r2d2}  & $\surd$ &  & 110M & 76.8 & 84.5 & 88.0  & & 78.9 & 83.5 & 86.0 \\
\hline
\textbf{YONO\scalebox{.7}[1.0]{$_{Reranking200}$}} &  & & 55M & \textbf{79.1} & \textbf{86.7} & 90.7 & & 82.1 & 86.0 & 88.1 \\
\textbf{YONO\scalebox{.7}[1.0]{$_{Reranking800}$}} &  & & 55M & \textbf{79.1} & 86.6 & \textbf{91.1} & & \textbf{82.3} & \textbf{86.4} & \textbf{88.7} \\
\hline
\end{tabular}
\caption{Recall@N results on Natural Questions and TriviaQA test sets. The best retrieval and reranking scores except larger models are indicated in bold. Reranking200/800 refer to reranking the 200/800 retrieved passages. The GAR$^+$ model uses a further 406M params for augmenting the query.}
\label{main_result_recall} 
\end{small}
\end{center}
\vskip -0.1in
\end{table*}

\subsection{Model Configurations}
We primarily compare our model with baselines that use 440M parameters in total. To get a single language model with 440M parameters, we initialize our model from the pre-trained T5-large \cite{t5} discarding final 18 decoder layers. This results in our model with 24 encoder and 6 decoder layers. The retrieval layer uses the first 12 encoder layers that uses 25\% fewer parameters than baselines' bi-encoders retrievers (165M vs 220M). Since our reranking layer works on the representations of the retrieval layer, we only allocate 4 encoder layers for reranking. The total number of parameters used for retrieval and reranking is 220M. The remaining 220M parameters are allocated for the reading layer. 

\subsection{Training Details}
At the first pre-training iteration, the model is trained with a batch of 800 \textit{question-passage-answer} triples for 100K steps. From the second iteration, we train the model for 1,250 steps per iteration using a batch with 64 \textit{question-passages-answer} triplets, where each triplet is packed with 100 retrieved passages. In total, we run 42 additional iterations after the first iteration for pre-training. 

After pre-training, the model is fine-tuned the same way as pre-training, except it is trained for 1 epoch at every iteration. 

The model is optimized with the Adam optimizer \cite{2015-kingma} with a learning rate $10^{-4}$. The first iteration takes 24 hours, and other iterations take around 5 hours each including MIPS indexing and passage refresh on 8 A100 GPUs. The penalty $\gamma$ for attention scores of the random in-batch negative passages is set to 5 in all our experiments.

We found that answer generation more easily over-fits compared to the retrieval during fine-tuning. To prevent this over-fitting, the model is once reinitialized from the pre-trained YONO model at the $6^{th}$ iteration after the model achieves acceptable recall on the downstream task.

\subsection{Datasets}
We evaluate our model with two standard open-domain question answering datasets, Natural Questions \cite{nq} and TriviaQA \cite{joshi2017triviaqa} following short answer subsets processed by \citet{orqa}. Our external knowledge base is built using the Wikipedia dump from Dec. 20, 2018, where articles are split into passages of 100 words without overlap which is the same as datasets used in \citet{dpr,fid,emdr2} for a fair comparison.

\section{Evaluation}
\subsection{Retrieval Performance}
\cref{main_result_recall} and \cref{main_result_em} show overall performance of our model and other baselines on Natural Questions and TriviaQA test sets. Our retrieval layer achieves the state-of-the-art recall@20/100 on Natural Questions regardless of model size even when compared with models with more than 4x model parameters. On TriviaQA, ours performs slightly worse than the state-of-the-art models, FiD-KD \cite{fidkd} and E2NR \cite{e2nr}, which use passage labels during training or 4x more parameters. Our approach achieves such performance without using passage labels making relevance for a larger range of applications that may not have these annotations. These results also do not use augmented data and as we show in \cref{emss_effect} it only gives slight improvements on the end-to-end performance.

\subsection{Reranking Performance} 
As shown in \cref{main_result_recall}, the reranking layer further improves the recall of our retrieved passages by 3.8 and 5.5 absolute recall@5 on NQ and TQA respectively when reranking 800 retrieved passages. This is 2.3 and 3.4 absolute point improvements over the previous state-of-the-art reranker model. Our model achieves these recall performances using only 55M parameters which is only half the size of the other reranker models. Similar to our retriever, our reranker does not require passage labels, unlike other rerankers. This improvement in recall when using the reranker persists even when reranking only 200 passages.

\begin{table}[t]
\begin{center}
\begin{small}
\setlength{\tabcolsep}{0.3em}
\begin{tabular}{lccc}
\hline
\textbf{Model} &  \textbf{\# Params} & \textbf{NQ}  & \textbf{TQA}\\ 
\hline
\multicolumn{4}{l}{\textit{Discriminative models}} \\
OrQA \cite{orqa} & 330M & 33.3  &  45.0 \\
REALM \cite{realm} & 330M & 40.4  & - \\
ANCE \cite{ance} & 330M & 46.0 & 57.5 \\
\hline
\multicolumn{4}{l}{\textit{Generative models}} \\
RAG \cite{rag} & 440M & 44.5 &  56.8  \\
FiD \cite{fid}& 440M & 48.2 & 65.0 \\
FiD-KD \scalebox{.9}[1.0]{\cite{fidkd}} & 440M & 49.6 & 68.8 \\
E2NR \cite{e2nr}& 440M & 45.9 & 56.3 \\
EMDR$^2$ \cite{emdr2}& 440M & 52.5 & 71.4 \\
\hline
\multicolumn{4}{l}{\textit{Larger models}} \\
E2NR \cite{e2nr} & 1.4B& 48.1 & 59.6 \\
FiD \cite{fid} & 990M & 51.4 & 67.6 \\
FiD-KD \scalebox{.9}[1.0]{\cite{fidkd}} & 990M & 53.7 & 72.1 \\
UnitedQA \cite{cheng-etal-2021-unitedqa} & 2.09B& 54.7 & 70.5 \\
R2-D2 \cite{r2d2} & 1.29B& 55.9 & 69.9 \\

\hline
\textbf{YONO$_{Retrieval}$} & 440M & 53.2 & 71.3  \\
\textbf{YONO$_{Reranking 200}$} & 440M & 53.2 & 71.5  \\
\textbf{YONO$_{Reranking 800}$} & 440M & \textbf{53.2} & \textbf{72.1}  \\
\hline
\end{tabular}
\caption{End-to-end Open QA Exact-Match results on Natural Questions and TriviaQA test sets. Our model uses top 100 retrieved or reranked passages to generate answers. The best EM scores except larger models are indicated in bold.}
\label{main_result_em}
\end{small}
\end{center}
\vskip -0.1in
\end{table}

\subsection{End-to-end Performance}
Our model achieves the best end-to-end performance among the models of the same size on NQ and irrespective of the model size on TQA as shown in \cref{main_result_em}. Our best scores improve EM scores by 0.7 points on both NQ and TQA respectively over the previously best performing model of the same size, EMDR$^2$ \cite{emdr2}. Using data augmentation further boosts these improvements on NQ by 0.3 as shown in \cref{tab:gqa}. Using reranking also improves the end-to-end scores on TQA by 0.8, a negligible improvement on NQ. We conjecture that this may be due to the higher recall of the retriever on NQ.

\section{Ablation Studies}

\subsection{The Reader Loss on Retrieval Performance}
\label{reader_loss}

\begin{table}[t]
\setlength{\tabcolsep}{0.20em}
\small
\centering
    \begin{tabular}{lccc}
        \hline
        \textbf{Loss} & \textbf{R@5} & \textbf{R@20} & \textbf{R@100} \\ 
        \hline
        $\mathcal{L}_{\scalebox{.9}[1.0]{\textit{retrieval}}}+\mathcal{L}_{\scalebox{.9}[1.0]{\textit{reader}}}$ & 28.8 & 48.1& 67.0 \\
        $\mathcal{L}_{\scalebox{.9}[1.0]{\textit{retrieval}}}$ & 18.0 & 32.1 & 49.7 \\
        \hline
        $\Delta$ & +10.8 {\tiny (60.0\%)}& +16.0 {\tiny (49.8\%)}& +18.7 {\tiny (34.8\%)}\\
        \hline
    \end{tabular}
\caption{Effect of reader's generation loss on zero shot retrieval performance after the first iteration of pre-training on Natural Questions development set.}
\label{tab:gen_loss_abl}
\end{table}

The retrieval layer is trained by signals from both retrieval and reader losses. While the retrieval loss directly trains the retrieval scores, the reader loss is also a useful indirect training signal for the retriever. This signal is a key advantage of our approach over similar works such as \citet{fidkd,emdr2}. We evaluate the performance gain from the additional generation loss at the first pre-training iteration as shown in \cref{tab:gen_loss_abl}. The model trained with both losses shows significant improvement over the model trained with only the retrieval loss. These relative gains are larger the fewer the number of retrieved passages. This result shows that reader's generation loss is very effective for training the retriever. We evaluate after the first iteration because the reader loss is necessary for training with multiple retrieved passages in the following iterations.

\subsection{Shared Representations on Reader Performance}
\label{shared_rep}

\begin{table}[ht]
\setlength{\tabcolsep}{0.20em}
\small
\centering
    \begin{tabular}{lcc}
        \hline
        \textbf{Model} & \textbf{Natural Questions} & \textbf{TriviaQA} \\
        \hline
        YONO Reader & 51.4 & 70.0 \\
        Stand-Alone Reader & 48.0 &  67.8 \\
        \hline
        $\Delta$ & +3.4 \tiny{(7.1\%)} &  +2.2 \tiny{(3.2\%)}\\
        \hline
    \end{tabular}
\caption{Effect of sharing of retrieval and reranking representations on exact match scores of reader models that use 220M parameters on NQ and TQA development sets.}
\label{tab:sep_reader}
\end{table}

Our reading layer uses 220M parameters but shares representations encoded by its preceding retrieval and reranking layers which use another 220M parameters. To measure gains of the shared representations, we compare our reader performance with that of a stand-alone reader model that uses 220M parameters that is the same as our reading layers. For a fair comparison, the stand-alone reader model is pre-trained and fine-tuned for the same amount of training tokens using the data retrieved by our YONO retriever. \cref{tab:sep_reader} shows that the reader model sharing representations outperforms the stand-alone reader by 7.1\% and 3.2\% on NQ and TQA respectively.

\begin{table}[t]
\setlength{\tabcolsep}{0.20em}
\small
\centering
    \begin{tabular}{lccccccc}
        \hline
                      & \multicolumn{3}{c}{\textbf{Natural Questions}} & & \multicolumn{3}{c}{\textbf{TriviaQA}} \\
         \cline{2-4} \cline{6-8} 
         \noalign{\vskip 0.7mm} 
        \textbf{Pre-training} & \textbf{\scalebox{.9}[1.0]{R@20}} & \textbf{\scalebox{.8}[1.0]{R@100}} & \textbf{EM} & & \textbf{\scalebox{.9}[1.0]{R@20}} & \textbf{\scalebox{.8}[1.0]{R@100}} & \textbf{EM} \\ 
        \hline
        \multicolumn{5}{l}{\textit{\textbf{Retrieval} }} \\
        No-pretrain     & 75.9 & 84.4 & 46.7 & & 60.5 & 77.7 & 55.1 \\
        MSS-P(1 iter.)     & 83.4 & 89.0 & 51.5 & & 80.4 & 85.9 & 68.4 \\
        MSS-P     & 85.2 & 90.2 & 53.2 & & \textbf{83.5} & 87.4 & \textbf{71.3} \\
        MSS-P+ASGen & \textbf{85.5} & \textbf{90.3} & \textbf{53.5} & & \textbf{83.5} & \textbf{87.5} & 70.9\\
        \hline
        \multicolumn{5}{l}{\textit{\textbf{Reranking 200} }} \\
        No-pretrain     & 81.4 & 85.9 & 46.8 & & 75.8 & 79.7& 57.4 \\
        MSS-P(1 iter.)     & 85.7 &  89.7 & 51.2 & & 84.0 & 86.6 & 69.1 \\
        MSS-P     & 86.7 & 90.7 & \textbf{53.2} & & 86.0 & 88.1 & \textbf{71.5} \\
        MSS-P+ASGen & \textbf{87.2} & \textbf{90.9} & \textbf{53.2} & & \textbf{86.2} & \textbf{88.2} & 71.2\\
        \hline
    \end{tabular}
\caption{Effect of MSS-P pre-training and further pre-training using augmented data on Natural Questions and TriviaQA test sets.}
\label{tab:gqa}
\end{table}

\subsection{Effectiveness of MSS-P pre-training}
\label{emss_effect}
To show the effectiveness of our MSS-P pre-training method, we evaluate this by fine-tuning our architecture without any pre-training using initial retrievals from DPR \cite{dpr} and fine-tuning after the first pre-training iteration. We also compare our pre-training to that of the additional data augmentation \cite{gar,pair,dprpaq}. We generate `question-answer' pairs from a Wikipedia dump using a question and answer generation model trained on the NQ dataset using the ASGen approach \cite{back-etal-2021-learning}. The model is further trained after the pre-training by this augmented data for 12 more iterations before fine-tuning. 

\cref{tab:gqa} shows retrieval, reranking, and reading performance on Natural Questions and TriviaQA test sets. Our MSS-P pre-training dramatically boosts the performance of our architecture by 4.8 and 13.3 EM points on NQ and TQA even with only the first pre-training iteration. Further iterations of our pre-training improve EM by 1.7 and 2.9 EM points. The further data augmentation pre-training improves performance on NQ consistently but only slightly, while the improvements on TQA are inconsistent, as the data was generated by the model trained on NQ dataset. These results clearly demonstrate that our simple self-supervised MSS-P pre-training is strong enough to compete favorably against sophisticated data augmentation approaches.

\section{Analysis}

\begin{figure}[t]

\centering
\footnotesize
\begin{tikzpicture}
\begin{axis}[
    name=axis1,
    xlabel={N passages},
    ylabel={Exact Match@N},
    height=5.5cm,
    xmin=-10, xmax=110,
    ymin=35, ymax=53,
    xtick={1,5,10,20,50,100},
    ytick={35,40,45,50,51},
    legend pos=south east,
    ymajorgrids=true,
    grid style=dashed,
    ylabel near ticks 
]

\addplot[color=red,mark=star,]
coordinates {(1,41.7)(5,48.5)(10,50.69)(20,51.01)(50,51.12)(100,51.1)};

\addplot[color=blue,mark=square,]
coordinates  {(1,36.7)(5,46.1)(10,48.30)(20,49.48)(50,50.5)(100,51.1)};

\legend{N reranked passages,N retrieved passages}

\end{axis}

\end{tikzpicture}

\caption{Exact Match scores for given N retrieved or reranked passages on NQ development set. Rerank EM scores are from reranking only 100 retrieved passages.}
\label{rerank_graph}
\end{figure}
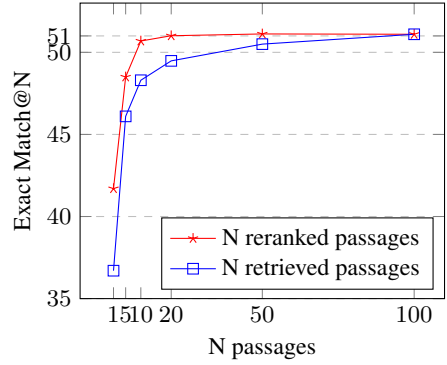

\subsection{Computational Efficiency of Reranking}
In many dense retrieval systems, a reranker is often omitted due to functional overlaps with the reader and computational overhead \cite{realm,rag,emdr2}. Thanks to the shared representations across the reader and reranker, our model can incorporate a reranking function without significantly more parameters or computation. By dropping irrelevant passages early at the reranking layer, we can achieve better computational efficiency. \cref{rerank_graph} shows exact match scores for given N retrieved or reranked passages. The model still achieves optimal EM performance with only the top 20 reranked passages reranked from 100 retrieved passages. Reranking 100 to 20 passages can reduce the inference computation by $27.4\%$ without pre-computed passage representations, and $54.0\%$ with pre-computed passage representations without losing end-to-end performance.

\subsection{Pre-training Efficiency}
\label{pre-train-overhead}
\begin{table}[t]
\setlength{\tabcolsep}{0.20em}
\small
\centering
    \begin{tabular}{lccccccc}
        \hline
        & \textbf{Input} & \multicolumn{2}{c}{\textbf{First Stage}} && \multicolumn{2}{c}{\textbf{Second Stage}} &  \textbf{Total}\\
        \cline{3-4}\cline{6-7}
        & \textbf{Len.} & \textbf{\scalebox{.9}[1.0]{Passages}} & \textbf{Steps} && \textbf{\scalebox{.9}[1.0]{Passages}} & \textbf{Steps} & \textbf{Tokens}\\
        \hline
        REALM & 288 & 4096 & 100K && 4096  & 200K & 352B \\
        EMDR$^2$ & 256 & 4096  & 100K && 3200 & 82K &  171B\\
        \hline
        \textbf{YONO} & 200 & 800 & 100K && 6400 & 52.5K & \textbf{83B}\\
        \hline
    \end{tabular}
\caption{Total pre-training tokens. The first and second stages of REALM and EMDR$^2$ are ICT and MSS pre-training respectively. }
\label{tab:tokens}
\end{table}

The number of pre-training tokens is an important metric to measure the efficiency of the pre-training objective. As shown in \cref{tab:tokens}, REALM \cite{realm} and EMDR$^2$ \cite{emdr2} use 352B and 171B tokens in total respectively. In contrast, our method uses only 83B tokens, which is 76.4\% and 51.5\% less than the training tokens used to train REALM and EMDR$^2$ respectively. Furthermore, the retrieval index is updated only 43 times during our pre-training, while EMDR$^2$ updates the index 164 times. This is a significant reduction of the computation overhead for pre-training.

\section{Related Works}

\paragraph{Neural Retriever Augmented Language Modeling (NRALM):}

Augmenting language models with neural retrieval has been shown to be very effective, such as by retrieving nearest neighbor words for LM tasks~\citep{knnlm, spalm} or Machine Translation ~\citep{knnmt}. \citet{wizardofwiki} proposed a decomposed transformer for conversation tasks, which enabled pre-computation of the external knowledge embeddings.

ORQA~\citep{orqa} proposed the ICT task to pre-train a decomposed retriever, and DPR~\citep{dpr} enhanced this approach with in-batch negatives and hard negatives to eliminate the pre-training. Synthetic Data Augmentation is also commonly used, such as in DPR-PAQ~\citep{dprpaq}, PAIR~\citep{pair}, \citet{rgpt}. Per-token embeddings or multiple embeddings were used in ColBERT~\citep{colbertqa}, ME-BERT~\citep{mebert}, \citet{inbatchkd}, \citet{densephrases}.

Similar to our approach of re-ranker on top of a shared retriever, PreTTR~\citep{prettr} pre-computed term representations for all documents, and used these to run only the upper layers of a transformer reranker model. Decoupled Transformer~\citep{decoupledtransformer} also shares the lower layers of a transformer encoder to serve as a reranker, using the upper layers as a reader and focuses on computationally efficient reranking. Our approach extends these approaches by also incorporating a retriever and a decoder in the model. 

\paragraph{E2E Optimization of NRALM:}
It is intractable to re-compute the embeddings of the knowledge for every weight update. REALM~\citep{realm} and ANCE~\citep{ance} proposed async index refresh to propagate updates to the index to yield better negatives. TAS~\citep{tas} and \citet{proqa} used clustering of embeddings for the same. RAG~\citep{rag} used DPR with BART generator to marginalize over generated tokens, which is back-propagated to the retriever. REALM++~\citep{realmplusplus} added a re-ranker to REALM. 

Similar to our work, \citet{kddconverse} and TREAD~\citep{tread} utilize BART and T5 reader's encoders as a retriever. In contrast to these methods, our work has a unified pre-training method to train all the components of the model. Furthermore, our model also has an integrated re-ranker, and the query and passage are cross-encoded for more expressive representations. 


\paragraph{Multi-passage Readers:}

Reading multiple passages at the same time is difficult, as concatenating multiple passages increases computation quadratically for transformers. \citet{knowledgegpt} reduced multiple passages and sentences to few via a knowledge selector, which were then concatenated and passed on to GPT \cite{radford2019language}. FiD~\citep{fid} concatenated the encoded representations of documents, which can then be attended by the decoder, achieving large performance gains. This approach was also applied in RocketQA~\citep{rocketqa}. UnitedQA~\citep{cheng-etal-2021-unitedqa} and R2D2~\citep{r2d2} combine results from an ensemble of extractive and generative readers, whereas PAQ~\citep{paq} directly retrieves answers with an FiD fallback. 

Similar to our work, both REALM~\citep{fidkd} and EMDR$^2$~\citep{emdr2} train the retriever with a signal from the reader. Unlike these approaches, our model has shared lower layers for more effective utilization of model parameters and better end-to-end gradient flow across the whole model. Furthermore, our training methodology results in propagating the answer generation loss of the retriever, which has a large effect on  performance as we show in \cref{tab:gen_loss_abl}. 



\section{Conclusion}
In this paper, we propose a novel language model architecture that embeds the retriever and the reranker as internal passage-wise attention mechanisms and a training method to effectively train this model. This singular model architecture efficiently uses model capacity by cascading and sharing the representations from retriever to reranker to the reader leading to better gradient flow for end-to-end training. We evaluate our model on Natural Questions and TriviaQA open datasets and for a fixed parameter budget, our model outperforms the previous state-of-the-art model by 1.0 and 0.7 exact match scores. We show detailed ablations and analyses of each component of our approach. Our future work is to conduct more experiments on various knowledge-intensive tasks and extend this model to match query and passage in multiple or hierarchical representation spaces.
\section*{Limitations}
\label{sec:limitation}

One caveat of sharing representation for multiple tasks like retrieval, reranking, and reading is that these show different over-fitting tendencies during fine-tuning where the training data is limited. We found that answer generation over-fits more easily compared to the retrieval. Answer generation relies on more expressive representation via cross attention, which may make it easier to memorize the output and hence make it more vulnerable to over-fitting. Furthermore, at the first fine-tuning iteration, the model is trained by zero-shot retrieval results from the pre-trained model that has a relatively low recall rate and can harm the answer generation training. To refresh the over-fitted answer generation parameters, and to start from training data with a high recall rate, we simply re-initialize the model with the pre-trained YONO model after a few fine-tuning iterations. However, we believe that this issue should be addressed carefully using a more sophisticated solution. We further discuss the over-fitting issue and effect of re-initialization in \cref{sec:reinit}.

\bibliography{emnlp2022}
\bibliographystyle{acl_natbib}

\appendix

\section*{Appendix}
\label{sec:appendix}

\section{Model Re-initialization during Fine-tuning}
\label{sec:reinit}

To overcome over-fitted reader parameters, we refresh the model parameter using the pre-trained YONO model at the fine-tuning iteration where the EM score starts to drop.

\begin{figure}[ht]

\centering
\footnotesize
\begin{tikzpicture}
\begin{axis}[
    name=axis3,
    xlabel={Iterations},
    ylabel={Exact Match},
    axis y line*=right,
    height=5.5cm,
    xmin=0, xmax=16,
    ymin=44, ymax=54,
    xtick={1,2,3,4,5,6,7,8,9,10,11,12,13,14,15},
    ytick={45,47,49,51,53},
    legend pos=south east,
    ymajorgrids=true,
    grid style=dashed,
    ylabel near ticks 
]
\addlegendimage{color=red,mark=star,}\addlegendentry{Recall@100}

\addplot[color=blue,mark=square,error bars/.cd, y dir=both,y explicit,]
coordinates  {
(1,45.26)+-(0.06,0.06)
(2,48.73)+-(0.07,0.07)
(3,50.21)+-(0.28,0.28)
(4,50.86)+-(0.21,0.21)
(5,50.55)+-(0.21,0.21)
};
\addplot[color=blue,mark=square,error bars/.cd, y dir=both,y explicit,]
coordinates  {
(6,47.20)+-(0.09,0.09)
(7,49.55)+-(0.09,0.09)
(8,50.35)+-(0.10,0.10)
(9,50.43)+-(0.19,0.19)
(10,51.03)+-(0.23,0.23)
(11,50.75)+-(0.19,0.19)
(12,50.77)+-(0.26,0.26)
(13,50.89)+-(0.28,0.28)
(14,50.87)+-(0.22,0.22)
(15,50.49)+-(0.23,0.23)
};\label{em100}

\addlegendentry{ExactMatch@100}

\end{axis}

\begin{axis}[
    height=5.5cm,
    xmin=0, xmax=16,
    ymin=84.5, ymax=89.5,
    ytick={85, 86, 87, 88, 89},
    hide x axis,
    axis y line*=left,
    ylabel={Recall},
    ylabel near ticks 
]

\addplot+[color=red,mark=star,error bars/.cd, y dir=both,y explicit,]
coordinates {
(1,86.00)+-(0.07,0.07)
(2,87.90)+-(0.03,0.03)
(3,88.50)+-(0.07,0.07)
(4,88.80)+-(0.01,0.01)
(5,88.90)+-(0.02,0.02)
} ;
\addplot[color=red,mark=star,error bars/.cd, y dir=both,y explicit,]
coordinates {
(6,87.94)+-(0.11,0.11)
(7,88.31)+-(0.05,0.05)
(8,88.73)+-(0.04,0.04)
(9,88.91)+-(0.08,0.08)
(10,88.94)+-(0.08,0.08)
(11,88.99)+-(0.11,0.11)
(12,89.06)+-(0.08,0.08)
(13,89.07)+-(0.05,0.05)
(14,89.09)+-(0.05,0.05)
(15,89.19)+-(0.06,0.06)
};

\addplot[color=black,dotted]
coordinates {(6,84)(6,90)}node [anchor=north, text=black, rotate=90,xshift=-2.5cm,yshift=0.4cm] {Model Reinitialization};

\end{axis}

\end{tikzpicture}

\caption{Average Recall and EM scores at each fine-tuning iteration with standard error bars from 3 runs on NQ development set. The model is once reinitialized at the $6^{th}$ iteration.}
\label{fig:model_reinit}
\end{figure}
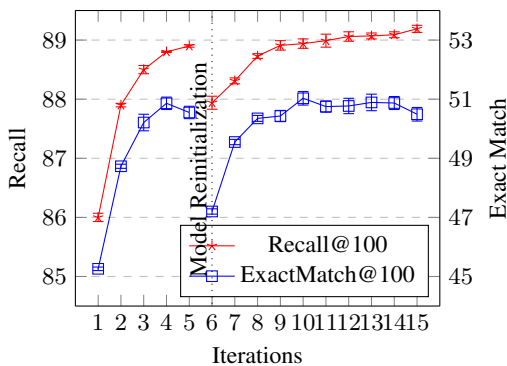

\cref{fig:model_reinit} shows retrieval and end-to-end performance at each fine-tuning iteration. The Exact Match score drops at the $5^{th}$ iteration while the retrieval score keeps increasing. After re-initializing the model before the $6^{th}$ iteration, the model re-starts with a higher recall and EM score. However, the EM score drops again from the $10^{th}$ iteration after achieving the best end-to-end performance, while the retrieval performance continues to improve. We leave further approaches for preventing over-fitting of our model such as freezing the model partially as future work.

\section{Effect of Pre-training Iterations on Retrieval Performance}
\begin{figure}[ht]

\centering
\footnotesize
\begin{tikzpicture}
\begin{axis}[
    name=axis2,
    xlabel={N passages},
    ylabel={Recall@N},
    height=6.5cm,
    xmin=-10, xmax=110,
    ymin=0, ymax=100,
    xtick={1,5,10,20,50,100},
    ytick={20,40,60,80},
    legend pos=south east,
    ymajorgrids=true,
    grid style=dashed,
    ylabel near ticks 
]

\addplot[color=purple,mark=diamond,]
coordinates {(1,58.8)(5,80.0)(10,84.1)(20,86.8)(50,89.0)(100,90.2)};

\addplot[color=red,mark=triangle,]
coordinates {(1,52.5)(5,75.4)(10,81.0)(20,84.7)(50,87.7)(100,89.3)};

\addplot[color=green,mark=star,]
coordinates {(1,16.1)(5,36.4)(10,46.11)(20,55.5)(50,65.6)(100,71.8)};

\addplot[color=blue,mark=square,]
coordinates  {(1,10.9)(5,28.8)(10,38.63)(20,48.1)(50,60.0)(100,67.0)};
    
\legend{F.T (Reranking800),F.T (Retrieval),P.T,P.T ($1^{st}$ iter.)}

\end{axis}
\end{tikzpicture}

\caption{Recall@N at each training stage on NQ development set. P.T denotes pre-training, F.T denotes fine-tuning.}
\label{recall_graph}
\end{figure}
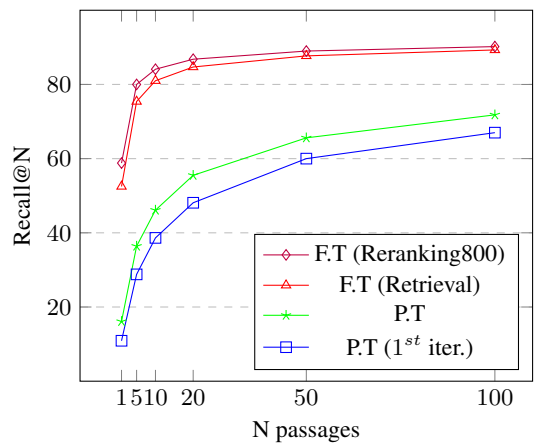

\cref{recall_graph} shows recall@N at each training stage across the pre-training and fine-tuning using Natural Question development set. The first iteration of the pre-training results in zero-shot recall@100 of $67.0\%$, which is further improved by additional pre-training iterations to $71.8\%$ recall@100.  These zero-shot recall scores enable us to fine-tune our model without passage labels resulting in state-of-the-art retrieval and reranking performance.

\section{Experiment Details}

On Table~\ref{model-hyperparam}, \ref{train-hyperparam}, and \ref{other-hyperparam}, we provide all training details and parameters used to conduct experiments on this paper. 

\begin{table}[ht]
\begin{center}
\begin{small}
\begin{tabular}{lcc}
\hline
\textbf{Parameters} & \multicolumn{2}{c}{\textbf{Values}} \\
\hline
Initial model & \multicolumn{2}{c}{\href{https://huggingface.co/t5-large}{T5-Large}} \\
Dimensions & \multicolumn{2}{c}{} \\
- Model & \multicolumn{2}{c}{1,024} \\
- Feed Forward & \multicolumn{2}{c}{4,096} \\
- Attention head & \multicolumn{2}{c}{64} \\
Attention head count & \multicolumn{2}{c}{16} \\
\# of layers and parameters& &  \\
- Total  & 24 enc + 6 dec & 440M \\
- Retrieval layer & 12 enc & 165M \\
- Reranking layer & 4 enc & 55M \\
- Reading layer & 8 enc + 6 dec & 220M\\
\hline
\end{tabular}
\caption{\label{model-hyperparam} Model Parameters.}
\end{small}
\end{center}
\vskip -0.1in
\end{table}

\begin{table}[ht]
\begin{center}
\begin{small}
\begin{tabular}{lc}
\hline
\textbf{Parameters} & \textbf{Values} \\
\hline
\textit{\textbf{Pre-training} }\\
Total iterations & 43 \\
Training tokens &  \\
- Total & 83B \\
- The $1^{st}$ iteration & 16B \\
- One iteration from $2^{nd}$ & 1.6B \\
- One batch & 160K \\
\hline
\textit{\textbf{Fine-tuning} }\\
Best scoring iteration &  NQ 10 / TQA 11 \\
Training tokens &  \\
- Total to the best iteration & NQ 15.8B / TQA 17.3B \\
- One iteration &  NQ 1.58B / TQA 1.58B \\
- One batch & 160K \\
\hline
\textit{\textbf{Optimization} }\\
Learning rate & $10^{-4}$ (fixed) \\
Drop-out & 0.1 \\
Precision & float32 \\
Gradient clipping & 1.0 \\
Gradient accumulation &  \\
- The $1^{st}$ pretraining iteration & None \\
- otherwise & 8 batches \\
\hline
\end{tabular}
\caption{\label{train-hyperparam} Training Parameters.}
\end{small}
\end{center}
\vskip -0.1in
\end{table}

\begin{table}[ht]
\begin{center}
\begin{small}
\begin{tabular}{lc}
\hline
\textbf{Parameters} & \textbf{Values} \\
\hline
\textit{\textbf{Text Inputs} }\\
Max question length & 40 \\
Max sequence length & 200 \\
\hline
\textit{\textbf{Retrieval Index} }\\
Dimension & 1,024\\
Precision & float32 \\
Index Size (21M passages) & 81GB \\
\hline
\end{tabular}
\caption{\label{other-hyperparam} Other Parameters.}
\end{small}
\end{center}
\vskip -0.1in
\end{table}

\section{Raw Values for Plots in Figures}
In Table~\ref{raw_fig2} and~\ref{raw_fig3}, we provide raw values for plots in Figure~\ref{rerank_graph} and \ref{recall_graph}.
\begin{table}[H]
\small
\centering
    \begin{tabular}{ccc}
        \hline
        \textbf{Passages} & \textbf{Retrived EM} & \textbf{Reranked EM} \\
        \hline
1 & 36.7 & 41.7 \\
5 & 46.1 & 48.5 \\
10 & 48.3 & 50.7 \\
20 & 49.5 & 51.0 \\
50 & 50.5 & 51.1 \\
100 & 51.1 & 51.1 \\
        \hline
    \end{tabular}
\caption{Raw Values for EM for \cref{rerank_graph}}
\label{raw_fig2}
\end{table}

\begin{table}[H]
\small
\centering
    \begin{tabular}{ccccc}
        \hline
        \textbf{Passages} & \textbf{FT \scalebox{.7}[1.0]{Rerank$_{800}$}} & \textbf{FT \scalebox{.7}[1.0]{Retrieval}} & \textbf{PT} & \textbf{PT \scalebox{.7}[1.0]{1 iter.}} \\
        \hline
1 & 58.8  & 52.5  & 16.1  & 10.9  \\
5 & 80.0  & 75.4  & 36.4  & 28.8  \\
10 & 84.1  & 81.0  & 46.1  & 38.6  \\
20 & 86.8  & 84.7  & 55.5  & 48.1  \\
50 & 89.0  & 87.7  & 65.6  & 60.0  \\
100 & 90.2  & 89.3  & 71.8  & 67.0  \\
        \hline
    \end{tabular}
\caption{Raw Values for Recall@N for \cref{recall_graph}}
\label{raw_fig3}
\end{table}

\section{Measures of Central Tendencies for Results}

To measure the sensitivity of our model to varying seeds, we run 3 fine-tunings of our model on NQ, and report the mean and standard errors on the development set below, as shown in \cref{fig:model_reinit}. The model training seems stable with little variation across runs. We did not run multiple instances of pre-training as it is computationally expensive.

\begin{table}[ht]
\small
\centering
    \begin{tabular}{ccc}
        \hline
        \textbf{Finetuning Iter.} & \textbf{Exact Match@100} & \textbf{Recall@100} \\
        \hline
1 & 45.3 $\pm$ 0.1  & 86.0 $\pm$ 0.1  \\
2 & 48.7 $\pm$ 0.1  & 87.9 $\pm$ 0.0  \\
3 & 50.2 $\pm$ 0.3  & 88.5 $\pm$ 0.1  \\
4 & 50.9 $\pm$ 0.2  & 88.8 $\pm$ 0.1  \\
5 & 50.6 $\pm$ 0.2  & 88.9 $\pm$ 0.1  \\
        \hline
6 & 47.2 $\pm$ 0.1  & 87.9 $\pm$ 0.1  \\
7 & 49.6 $\pm$ 0.1  & 88.3 $\pm$ 0.1  \\
8 & 50.4 $\pm$ 0.1  & 88.7 $\pm$ 0.1  \\
9 & 50.4 $\pm$ 0.2  & 88.9 $\pm$ 0.1  \\
10 & 51.0 $\pm$ 0.2  & 88.9 $\pm$ 0.1  \\
11 & 50.8 $\pm$ 0.2  & 89.0 $\pm$ 0.1  \\
12 & 50.8 $\pm$ 0.3  & 89.1 $\pm$ 0.1  \\
13 & 50.9 $\pm$ 0.3  & 89.1 $\pm$ 0.1  \\
14 & 50.9 $\pm$ 0.2  & 89.1 $\pm$ 0.1  \\
15 & 50.5 $\pm$ 0.2  & 89.2 $\pm$ 0.1  \\
        \hline
    \end{tabular}
\caption{Raw values for Central Tendency and Standard Error on NQ development set for 3 finetuning runs of our model, as shown in \cref{fig:model_reinit}}
\end{table}

\section{Links to Source Code and Datasets}

The source code is based on the original implementation of FiD~\citep{fid}, which can be found at \href{https://github.com/facebookresearch/FiD}{their Github}. 

Data for the Wikipedia dump, Natural Questions, and TriviaQA can also be downloaded from FiD's github using  \href{https://github.com/facebookresearch/FiD/blob/25ed1ff0fe0288b80fb5e9e5de8d6346b94b8d48/get-data.sh}{this script}.

\section{Evaluation Metrics and Scripts}

The evaluation script is based on the original FiD \href{https://github.com/facebookresearch/FiD/blob/25ed1ff0fe0288b80fb5e9e5de8d6346b94b8d48/src/evaluation.py}{script}.

Exact Match - This is the average across all examples of the per-example exact match score, which is 0 or 1 if all the words in the generated answer exactly match the annotated answer after unicode normalization by lower-casing, removing punctuation and spaces. 

Recall@N - Recall@N measures the percentage of examples for which atleast one the top-N passages contains a span that matches the annotated answer as in Exact Match above.

\section{Dataset Statistics}
Table~\ref{tab_dataset}
provides the statistics of our evaluation datasets. 

\begin{table}[h!]
\begin{center}
\begin{tabular}{lccc}
\hline
\textbf{Dataset} & \textbf{\# Train} & \textbf{\# Dev} & \textbf{\# Test} \\ 
\hline
Natural Questions & 79K & 8.8K & 3.6K   \\
TriviaQA & 79K & 8.8K & 11K  \\
\hline
\end{tabular}

\end{center}
\caption{Dataset Statistics}
\label{tab_dataset}
\end{table}

\section{Computing Infrastructure}
GPU model - 8x Nvidia A100 80 GB. CPU Model - 2x AMD EPYC 7543 32-Core Processor. RAM - 1000GB. PyTroch version - 1.8.0+cu111. Huggingface Transformers version - 3.0.2.

\end{document}